\documentclass{article}
\usepackage{spconf,amsmath,graphicx}
\usepackage{amssymb}
\usepackage{esvect}
\usepackage{color}
\usepackage{adjustbox}
\usepackage{tabu}
\usepackage{booktabs}

\usepackage{enumitem}
\setlist{nosep, leftmargin=14pt}

\usepackage{mwe} 

\title{RECIST-Net: Lesion Detection via Grouping Keypoints on RECIST-based Annotation}
\name{\parbox{\linewidth}{\centering Cong Xie\textsuperscript{1,2}\sthanks{C. Xie contributed to this work during an internship at Tencent.}, Shilei Cao\textsuperscript{2}, Dong Wei\textsuperscript{2}, Hongyu Zhou\textsuperscript{3}, Kai Ma\textsuperscript{2}, Xianli Zhang\textsuperscript{2,4}, Buyue Qian\textsuperscript{4},\\Liansheng Wang\textsuperscript{1}\sthanks{Correspondence: lswang@xmu.edu.cn}, Yefeng Zheng\textsuperscript{2}}}
\address{\textsuperscript{1}Xiamen University; \textsuperscript{2}Tencent Jarvis Lab; \textsuperscript{3}The University of Hong Kong; \textsuperscript{4}Xi'an Jiaotong University}
\begin{document}
%
\maketitle
\begin{abstract}
Universal lesion detection in computed tomography (CT) images is an important yet challenging task due to the large variations in lesion type, size, shape, and appearance. Considering that data in clinical routine (such as the DeepLesion dataset) are usually annotated with a long and a short diameter according to the standard of Response Evaluation Criteria in Solid Tumors (RECIST) diameters, we propose RECIST-Net, a new approach to lesion detection in which the four extreme points and center point of the RECIST diameters are detected. By detecting a lesion as keypoints, we provide a more conceptually straightforward formulation for detection, and overcome several drawbacks (\textit{e.g.,} requiring extensive effort in designing data-appropriate anchors and losing shape information) of existing bounding-box-based methods while exploring a single-task, one-stage approach compared to other RECIST-based approaches. Experiments show that RECIST-Net achieves a sensitivity of 92.49\% at four false positives per image, outperforming other recent methods including those using multi-task learning.
\end{abstract}
\section{Introduction}
\begin{figure}[!t]
\center
\includegraphics[width=1.0\columnwidth, trim=0 27 0 0, clip]{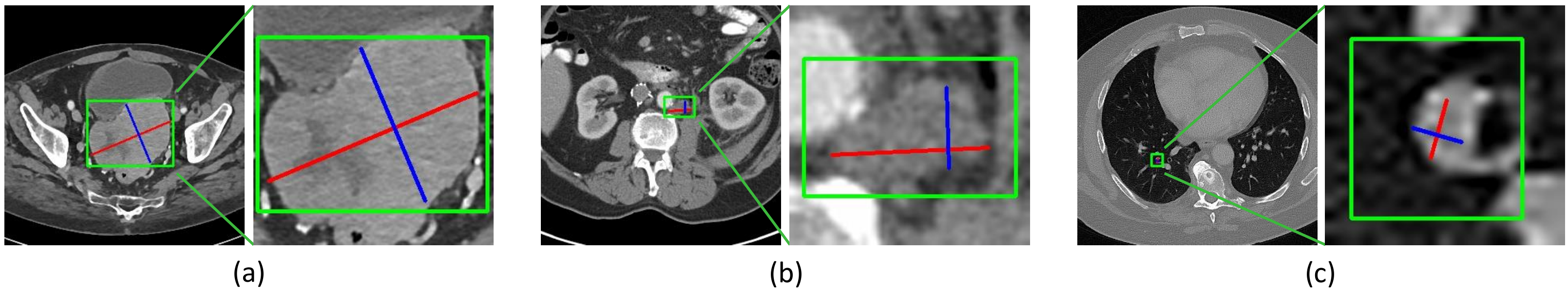}
\caption{Three example lesions of the DeepLesion \cite{yan2018deeplesion} dataset annotated with the RECIST diameters (red for the long diameters and blue for short). The bounding boxes are computed to enclose the lesion measurements with a 5-pixel padding in each direction.
Left to right: large to small lesions.}
\label{fig_illustration}
\end{figure}

Lesion detection plays an important role in computer-aided detection/diagnosis (CAD) systems. Early algorithms generally focused on one or few particular lesion types. To promote the development of universal lesion detection algorithms, Yan \textit{et al.}~\cite{yan2018deeplesion} built a large-scale dataset comprising lesions of multiple categories named DeepLesion. The DeepLesion dataset was annotated with the Response Evaluation Criteria in Solid Tumors (RECIST) diameters, which is one of the most frequently used ways of recording clinically meaningful findings in clinical routine by radiologists, due to its prevailing adoption for cancer patient monitoring. As part of the RECIST guidelines, the lesion
diameters include two lines, with the first measuring the longest diameter of the lesion and the second indicating the longest perpendicular diameter to the first in the plane of measurement (see Fig. \ref{fig_illustration} for examples). A bounding box is also provided for each lesion in DeepLesion, which is computed to enclose the diameter measurement with a 5-pixel padding in each direction (\textit{i.e.,} left, top, right, and bottom). Using the DeepLesion dataset, various methods \cite{yan20183d,shao2019attentive,tao2019improving,li2019mvp,yan2019mulan} have been proposed and advanced the state of the art for universal lesion detection. Although these methods yielded promising results, lesion detection using bounding boxes suffers from three prominent drawbacks.
First, the bounding box may not be a good representation for lesions.
Second, the process of tuning anchor boxes 
can be laborious.
Third, the bounding box is not the clinical standard for measuring lesion sizes.

Few works attempted to overcome one or several of the three drawbacks described above.
Tang \textit{et al.}~\cite{tang2019uldor} constructed a pseudo ellipse mask from the RECIST annotation for each lesion, and adopted Mask R-CNN \cite{he2017mask} for predicting the pseudo mask.
Zlocha \textit{et al.}~\cite{zlocha2019improving} improved quality of the pseudo mask with GrabCut \cite{rother2004grabcut}, and employed multi-scale dense supervision by a pseudo segmentation task to aid the detection task.
Two main drawbacks of these two works were that the pseudo mask was often inaccurate in segmenting the lesion, and extra overhead was incurred to learn the pseudo mask.
Different from these works, a noteworthy work \cite{tang2018semi} proposed to directly predict the endpoints of the RECIST diameters, which is (as far as we know) the first work that modeled these characteristics of the RECIST annotations.
However, the method was semi-automatic where the region of interest must be provided as a prerequisite, and the process was quite intricate where an extra spatial transformer network was employed to learn the transformation for unifying the lesion orientation.


\begin{figure*}[!t]
\center
\includegraphics[width=0.9\textwidth]{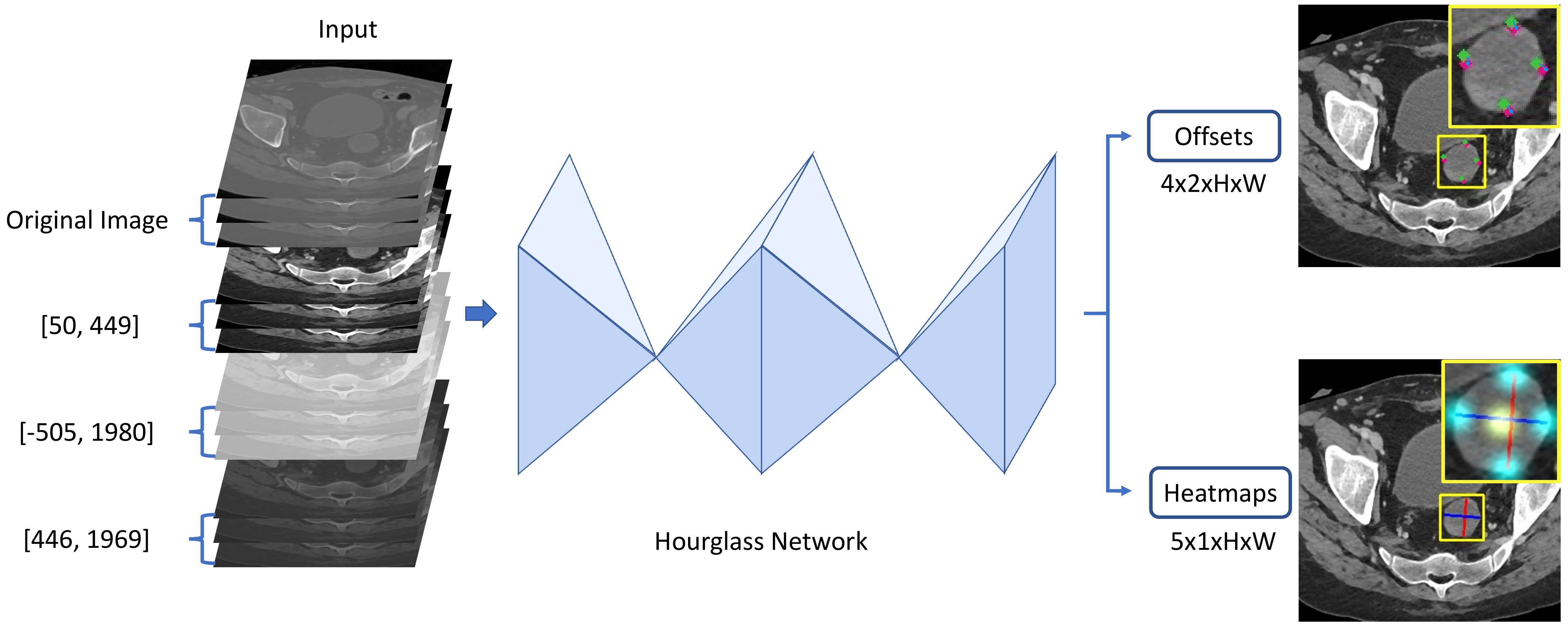}
\caption{An overview of the proposed RECIST-Net for detecting the extreme and center points of the RECIST diameters.}
\label{fig_framework}
\end{figure*}

In this paper, we propose a conceptually straightforward network named RECIST-Net to detect the four extreme points (\textit{i.e.}, top-most, left-most, bottom-most, right-most) and the center point of the RECIST diameters, overcoming all the limitations mentioned above.
The extreme and center points can characterize the lesions.
To learn these keypoints, we borrow ideas from the ExtremeNet \cite{zhou2019bottom} and employ an HourglassNet \cite{newell2016stacked} to regress the heatmaps of the extreme and center points by treating the task as a keypoint detection problem.
For testing, we propose a purely geometry-based grouping strategy to produce a bounding box for each prediction.
We evaluate our method on the DeepLesion dataset \cite{yan2018deeplesion}, and achieve a sensitivity of 92.49\% at four false positives per image, outperforming all competing methods including methods using multi-task learning \cite{yan2019mulan,li2019mvp,zlocha2019improving}.

\section{Methodology}
In this section, we present the details of the proposed RECIST-Net for detection of the extreme and center points of the RECIST diameters. An overview of the RECIST-Net architecture is presented in Fig. \ref{fig_framework}. We firstly briefly introduce the Hourglass backbone that we use. Then, we present the design of our detection head for learning the extreme and center points of RECIST diameters. Lastly, a geometry-based grouping strategy is explored to produce a bounding box for each detection.

\subsection{Hourglass Backbone} \label{hourglass}
Our RECIST-Net adopts the HourglassNet \cite{newell2016stacked} as backbone to detect the extreme and center points of the RECIST diameters.
For the input, considering that neighboring slices are important to providing contextual information for differentiating lesions from non-lesions, we group three consecutive axial slices of a CT volume into a 3-channel image.
In addition, Li \textit{et al.} \cite{li2019mvp} demonstrated that CT images with different window levels and widths could improve the performance in detecting subtle lesions and reducing false positives (FPs).
Being inspired, we stack images with three different configurations of window level and width to the original image as input.

\subsection{Learning Extreme and Center Points in RECIST Diameters} \label{extreme_center}
In this work, we grasp the central concepts of CornerNet \cite{law2018cornernet} and ExtremeNet \cite{zhou2019bottom}. Given four extreme and one center points, we regress a heatmap $\tilde{Y} \in [0, 1]^{H \times W}$ of width $W$ and height $H$ for each keypoint.
The training is guided by a multi-peak Gaussian heatmap $Y \in [0, 1]^{H \times W}$, where each keypoint defines the center of a Gaussian kernel and the standard deviation is set proportional to the object size \cite{law2018cornernet,zhou2019bottom}.
In order to balance the positive and negative locations, a modified focal loss is adopted for training, as in \cite{law2018cornernet,zhou2019bottom}:
\begin{equation}
\begin{aligned}
&\mathcal{L}_{det} = \\
&- \frac{1}{N} \sum_{i=1}^{H}\sum_{j=1}^{W}
\begin{cases}
(1-\tilde{Y}_{ij})^{\alpha}\mathrm{log}(\tilde{Y}_{ij}), &\mathrm{if} \text{ } Y_{ij}=1, \\
(1-\tilde{Y}_{ij})^{\beta}\tilde{Y}_{ij}^{\alpha}\mathrm{log}(1-\tilde{Y}_{ij}),  &\mathrm{otherwise},
\end{cases}
\end{aligned}
\end{equation}
where $\alpha$ and $\beta$ are hyper-parameters and fixed to $\alpha=2$ and $\beta=4$ during training, and $N$ is the number of objects in the image.

Similar to \cite{law2018cornernet,zhou2019bottom}, we additionally regress the keypoint offset
$\vartriangle_\mathrm{off}$ for each extreme point to recover part of the information lost in the down-sampling process of the HourglassNet \cite{newell2016stacked}.
We regress the offset maps with the smooth $\mathrm{L1}$ loss
$SL1$ on locations of the ground truth extreme point as:
\begin{equation}\label{eq:off}
\mathcal{L}_\mathrm{off}=\frac{1}{N}\sum_{k=1}^NSL1(\vartriangle_\mathrm{off}, \vv{\mathit{c}}/s - \lfloor \vv{\mathit{c}}/s \rfloor ),
\end{equation}where $s$ is the down-sampling factor in HourglassNet ($s=4$ in our case), and $\vv{\mathit{c}}$ is the coordinate of the estimated extreme point. Note that we omit the indexing of $k$ in the $SL1$ of Eq. (\ref{eq:off}) for convenience. There is no offset prediction for the center point.

\subsection{Grouping Strategy for Inference} \label{grouping_strategy}
Next, we present the strategy of how to group the prediction heatmaps into detections in a purely geometric manner for inference.
We simplify the grouping strategy used in ExtremeNet \cite{zhou2019bottom} into three big steps as described below.

First, for each heatmap of the extreme points, we extract local peaks with a max-pooling operation of kernel size $3\times3$ to suppress similar scores in neighborhood windows (named ExtractPeak in ExtremeNet \cite{zhou2019bottom}).
After the suppression, we preserve only the top $K_1$ positions with scores greater than a threshold $\tau_e$.
Second, given four extreme points denoted by $t, b, r, l$ in the corresponding heatmaps, their geometric center is calculated by $c=(\frac{\hat{x}^{(l)}+\hat{x}^{(r)}}{2}, \frac{\hat{y}^{(t)}+\hat{y}^{(b)}}{2})$.
If this center is predicted with a high response in the center-point heatmap, \textit{i.e.}, greater than a threshold $\tau_c$, we consider the extreme points as a valid candidate detection.
In this paper $\tau_e=\tau_c=0.1$.
We then iterate over all possible combinations of the remaining peak positions $t, b, r, l$ in a brute force manner (though with a runtime of $O(n^4)$, where $n\leq K_1$ is the number of preserved extreme points in corresponding heatmaps, it can be accelerated on a GPU).
A combination score is computed by adding up the scores of each quadruple of extreme points and twice the score of the corresponding center point, and the top $K_2$ candidate combinations are preserved as the initial prediction results for detection.
The settings of $K_1=40$ and $K_2=100$ in \cite{zhou2019bottom} are adopted in this paper.
Third, we refine the coordinates of the initial prediction results by adding an offset predicted at the corresponding location of the offset map to each predicted extreme point.
For a fair comparison with other methods which have been evaluated against bounding boxes, we generate a tight bounding box enclosing each grouped quadruple of detected extreme points.
Different from ExtremeNet \cite{zhou2019bottom}, which additionally employed a multi-scale augmentation for inference, we only use the flip augmentation for computational efficiency.
Lastly, a Soft-NMS is employed to filter all augmented detection results.

\section{Experiments}
\subsection{Dataset}
The DeepLesion dataset~\cite{yan2018deeplesion} is used for experiments, with the official split (70\% for training, 15\% for validation, and 15\% for test). As in \cite{yan20183d}, 35 noisy lesion annotations are removed. We evaluate performance of the methods on the official test set.
Following the general practice in previous works \cite{yan2019mulan, li2019mvp, zlocha2019improving, shao2019attentive, zhang20193d, tao2019improving, wang2019volumetric, wang2019towards, yan20183d}, we report the sensitivity at various FPs (0.5, 1, 2, 3, 4) per scan as the evaluation metric.

\subsection{Implementation}
During training, we set the input resolution of the axial slices to $511\times511$ in pixels, and output resolution to $128\times128$ in pixels.
To alleviate the overfitting problem, we use three data augmentation methods: random flipping horizontally and vertically, random scaling between 0.6 and 1.4, and random cropping.
In the test phase, the input axial image is resized to a fixed size of $639\times639$ in pixels.
The predicted bounding box coordinates are enlarged with a 5-pixel padding as done with the ground truth for computing sensitivity. The multi-view input is consistent with the window settings in \cite{li2019mvp}. We initialize our network using weights of ExtremeNet \cite{zhou2019bottom} trained on COCO. The network is optimized with Adam with a learning rate of $2.5 \times 10^{-4}$. We train our RECIST-Net on two NVIDIA GeForce RTX 2080 Ti GPUs with a batch size of 11 for 55,550 iterations.

\subsection{Comparison with State-of-The-Art Methods}
\begin{figure}[!t]
\center
\includegraphics[width=1.\columnwidth]{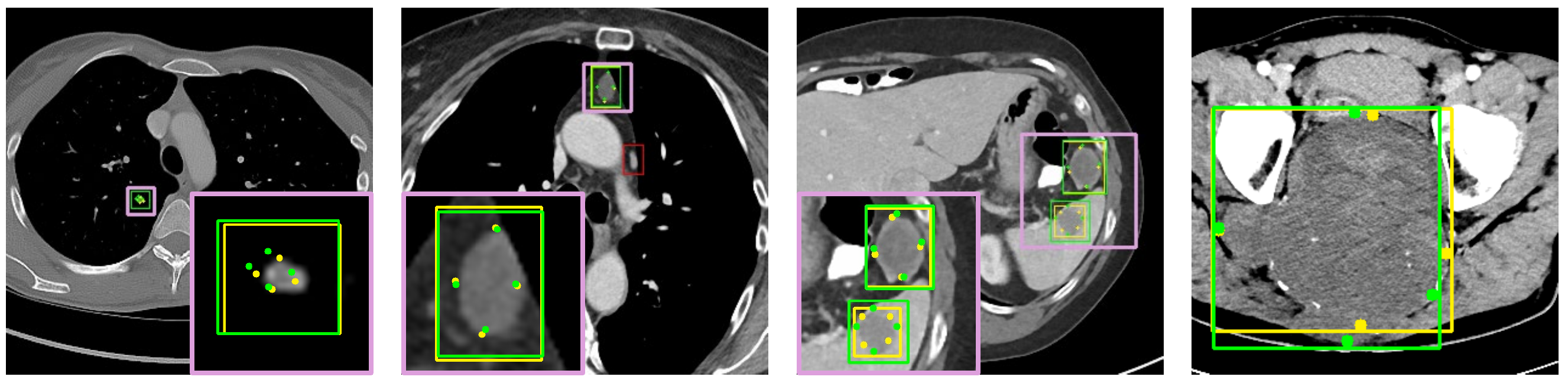}
\caption{Visual results for lesion detection at 0.5 FP rate using RECIST-Net. Yellow boxes/points are ground truth, green are true positives, red are false positives, and pink boxes on the bottom corners show enlarged views.}
\label{fig_visual_result}
\end{figure}

We show the results in Table \ref{tab_sota}. From the table, we can observe that our RECIST-Net outperforms all competing methods, including those using multi-task learning \cite{yan2019mulan,li2019mvp,tang2019uldor,zlocha2019improving}. The superior performance is attributed to the conceptually straightforward RECIST-based formulation for detection, which has not been exploited in previous works. Visual examples of the detected lesions on official test images are shown in Fig. \ref{fig_visual_result}. The probability threshold is set to 0.32 yielding 0.5 FP per image. We can observe that lesions of varying size, appearance, and type are localized accurately.

\begin{table}[!t]
\large
\centering
\setlength{\tabcolsep}{2.7mm}
\caption{Sensitivity (\%) at different false positives (FPs) per scan on the test set of the DeepLesion dataset \cite{yan2018deeplesion}.
Values for methods in comparison were reported in cited references based on the same train/validation/test split of the dataset.
Note that MULAN~\cite{yan2019mulan} used extra tag supervision, and MVP-Net~\cite{li2019mvp} used the same multi-view input as we do.}\label{tab_sota}
\begin{adjustbox}{width=1 \linewidth}
\begin{tabular}{lccccc}
\toprule
FPs per scan                                       & 0.5   & 1     & 2     & 3     & 4     \\ \midrule \midrule
3DCE, 27 slices~\cite{yan20183d}                    & 62.48 & 73.37 & 80.70 & -     & 85.65  \\
Faster-RCNN + DA~\cite{wang2019towards}               & -     & -     & -     & -     & 87.29   \\
Deformable Faster-RCNN + VA~\cite{wang2019volumetric} & 69.1  & 77.9  & 83.8  & -     & -    \\
3DCE + CS\_Att, 21 slices~\cite{tao2019improving}    & 71.4  & 78.5  & 84.0  & -     & 87.6  \\
Anchor-Free RPN~\cite{zhang20193d}                  & 68.73 & 77.10 & 83.54 & -     & 88.12 \\
FPN + MSB (weights sharing)~\cite{shao2019attentive}  & 67.0  & 76.8  & 83.7  & -     & 89.0  \\
Improved RetinaNet~\cite{zlocha2019improving}       & 72.15 & 80.07 & 86.40 & -     & 90.77 \\
MVP-Net, 9 slices~\cite{li2019mvp}                  & 73.83 & 81.82 & 87.60 & 89.57 & 91.30 \\
MULAN~\cite{yan2019mulan}                           & 76.12 & 83.69 & 88.76 & -     & 92.30 \\
\midrule
RECIST-Net (original image, 3 slices)                             & 74.33 & 81.80 & 87.66 & 90.02 & 90.68 \\
RECIST-Net (multi-view input)            & \textbf{76.14} & \textbf{83.71} & \textbf{89.62} & \textbf{91.69} & \textbf{92.49} \\
\bottomrule
\end{tabular}
\end{adjustbox}
\end{table}

We further analyze the detection performance on different lesion types and image properties on the official test set according to three criteria: 1) lesion type, 2) lesion diameter, and 3) slice interval of the CT scans.
We show the results per lesion type criterion in Table~\ref{tab_lesion_type} and those per other two criteria in Table \ref{tab_lesion_size}.
As shown, our method achieves the best performances on all metrics, except for the lesions with diameters greater than 30 mm, for which it achieves the second best.
This is because, for larger objects, the center response map \cite{zhou2019bottom} may not be accurate enough to perform well, as a shift of a few pixels might miss a detection and result in a false-negative \cite{zhou2019bottom}.

\begin{table}[!t]
 \centering
 \caption{Sensitivity (\%) at four FPs per scan on the test set of the DeepLesion dataset \cite{yan2018deeplesion}. We report the results on eight types of lesions as 3DCE \cite{yan20183d}, including lung (LU), mediastinum (ME), liver (LV), soft tissue (ST), pelvis (PV), abdomen (AB), kidney (KD), and bone (BN).
 Corresponding results reported in the literature are included for comparison.} \label{tab_lesion_type}
 \begin{adjustbox}{width=1.\linewidth}
 \begin{tabular}{l@{\hspace*{0.5em}}c@{\hspace*{0.5em}}c@{\hspace*{0.5em}}c@{\hspace*{0.5em}}c@{\hspace*{0.5em}}c@{\hspace*{0.5em}}c@{\hspace*{0.5em}}c@{\hspace*{0.5em}}c}
 \toprule
 Lesion Types                                                 & LU    & ME    & LV    & ST    & PV    & AB    & KD    & BN \\ \midrule \midrule
 3DCE, 27 slices~\cite{yan20183d}                 & 89.00 & 88.00 & 90.00 & 74.00 & 84.00 & 84.00 & 82.00 & 75.00 \\
 3DCE\_CS\_Att, 15 slices~\cite{tao2019improving} & 92.00 & 88.50 & 91.40 & 80.30 & 85.00 & 84.40 & 84.30 & 75.00 \\
 Anchor-Free RPN~\cite{zhang20193d}               & 93.00 & 88.00 & 91.00 & 85.00 & 86.00 & 83.00 & 80.00 & 65.00 \\ \midrule
 RECIST-Net                                             & \textbf{94.36} & \textbf{94.33} & \textbf{94.29} & \textbf{86.85} & \textbf{89.73} & \textbf{90.01} & \textbf{93.56} & \textbf{87.04} \\
 \bottomrule
 \end{tabular}
 \end{adjustbox}
 \end{table}

\begin{table}[!t]
\centering
\large
\setlength{\tabcolsep}{2.3mm}
\caption{Sensitivity (\%) at four FPs per scan on the test set of DeepLesion \cite{yan2018deeplesion}. We report results with different sizes and slice intervals of CT scans as 3DCE \cite{yan20183d}.
Corresponding results reported in the literature are included for comparison.}\label{tab_lesion_size}
\begin{adjustbox}{width=1 \linewidth}
\begin{tabular}{lccccc}
\toprule
                                                   & \multicolumn{3}{c}{Lesion diameter (mm)}      & \multicolumn{2}{c}{Slice interval (mm)} \\
                                                    & \textless{}10 & 10$\sim$30 & \textgreater{}30 & \textless{}2.5    & \textgreater{}2.5   \\ \midrule \midrule
3DCE, 27 slices~\cite{yan20183d}                    & 80.00 & 87.00 & 84.00 & 86.00 & 86.00 \\
Anchor-Free RPN~\cite{zhang20193d}                  & 83.00 & 87.00 & 88.00 & -     & -     \\
3DCE\_CS\_Att, 15 slices~\cite{tao2019improving}    & 82.30 & 90.00 & 85.00 & 87.60 & 87.60 \\
FPN+MSB (weights sharing)~\cite{shao2019attentive}  & 86.00 & 91.00 & -     & -     & -     \\
Improved RetinaNet~\cite{zlocha2019improving}       & 88.35 & 91.73 & \textbf{93.02} & -     & -     \\
\midrule
RECIST-Net                                                & \textbf{90.69} & \textbf{93.67} & 90.99 & \textbf{93.28} & \textbf{91.75} \\ \bottomrule
\end{tabular}
\end{adjustbox}
\end{table}

\subsection{Ablation Study}
We conduct an ablation study with respect to the ExtractPeak grouping strategy, Soft-NMS for post-processing, flip augmentation (FlipAug) for test time augmentation, and multi-view input, to identify how much these components contribute to the performance.
The results are shown in Table \ref{tab_ablation_study}.
We can observe that ExtractPeak and Soft-NMS bring the most improvements to RECIST-Net, both of which are used to suppress similar scores in neighborhood windows.
The result confirms the core status of the non-maximum-suppression-like methods in detection algorithms.
It is worth noting that with flip augmentation, a 7.66\% improvement is achieved in sensitivity at four FPs.
This may imply that detection with different orientations can provide complementary information.
With multi-view input, a further 1.81\% improvement is achieved.
This is reasonable, since, different window levels and widths are used for the reading of CT scans of different body parts in clinical practice, and hence a multi-view setting can boost the performance of lesion detection across the body.


\begin{table}[!t]
\centering
\large
\setlength{\tabcolsep}{1.1mm}
\caption{Ablation study on building components of RECIST-Net, where it is incrementally built up by adding to the baseline model (row (a)) one component at a time (rows (b)--(e)).
Sensitivity (\%) at different FPs per scan are reported.} \label{tab_ablation_study}
\begin{adjustbox}{width=1. \linewidth}
\begin{tabular}{ccccccccccc}
\toprule
& ExtractPeak & Soft-NMS     & FlipAug & Multi-View & 0.5   & 1     & 2     & 3     & 4     \\ \midrule
(a) &              &              &           &        & 13.80 & 19.97 & 28.79 & 34.32 & 38.42 \\
(b) & \checkmark   &              &      &             & 56.76 & 64.13 & 70.52 & 74.04 & 76.34 \\
(c) & \checkmark   & \checkmark   &          &         & 70.87 & 78.68 & 82.90 & 83.00 & 83.02 \\
(d) & \checkmark   & \checkmark   & \checkmark &        & 74.33 & 81.80 & 87.66 & 90.02 & 90.68 \\ \midrule
(e) & \checkmark   & \checkmark   & \checkmark & \checkmark      & \textbf{76.14} & \textbf{83.71} & \textbf{89.62} & \textbf{91.69} & \textbf{92.49} \\
\bottomrule
\end{tabular}
\end{adjustbox}
\end{table}


\section{Conclusion}
In this paper, we presented a formulation for universal lesion detection which was implemented with RECIST-Net.
The RECIST-Net detected four extreme points (\textit{i.e.,} top-most, left-most, bottom-most, right-most) and one center point of a lesion in a keypoint detection way. We hope this work would inspire researchers to develop more methods that are friendly to the way of annotation.

\section{Compliance with Ethical Standards}
This research study was conducted retrospectively using human subject data made available in open access by~\cite{yan2018deeplesion}.
Ethical approval was not required as confirmed by the license attached with the open access data.

\section{Acknowledgments}
This work was supported by National Natural Science Foundation of China (Grant No. 61671399) and the Fundamental Research Funds for the Central Universities (Grant No. 20720190012).
Shilei Cao, Dong Wei, Kai Ma, and Yefeng Zheng are employees of Tencent.
The authors have no relevant financial or non-financial interest to disclose.

\bibliographystyle{IEEEbib}

\begin{thebibliography}{10}

\bibitem{yan2018deeplesion}
Ke~Yan, Xiaosong Wang, Le~Lu, and Ronald~M. Summers,
\newblock ``{DeepLesion}: {A}utomated mining of large-scale lesion annotations
  and universal lesion detection with deep learning,''
\newblock {\em J. of Med. Imaging}, vol. 5, no. 3, pp. 036501, 2018.

\bibitem{yan20183d}
Ke~Yan, Mohammadhadi Bagheri, and Ronald~M. Summers,
\newblock ``{3D} context enhanced region-based convolutional neural network for
  end-to-end lesion detection,''
\newblock in {\em MICCAI}. Springer, 2018, pp. 511--519.

\bibitem{shao2019attentive}
Qingbin Shao, Lijun Gong, Kai Ma, Hualuo Liu, and Yefeng Zheng,
\newblock ``Attentive {CT} lesion detection using deep pyramid inference with
  multi-scale booster,''
\newblock in {\em MICCAI}. Springer, 2019, pp. 301--309.

\bibitem{tao2019improving}
Qingyi Tao, Zongyuan Ge, Jianfei Cai, Jianxiong Yin, and Simon See,
\newblock ``Improving deep lesion detection using {3D} contextual and spatial
  attention,''
\newblock in {\em MICCAI}. Springer, 2019, pp. 185--193.

\bibitem{li2019mvp}
Zihao Li, Shu Zhang, Junge Zhang, Kaiqi Huang, Yizhou Wang, and Yizhou Yu,
\newblock ``{MVP-Net: M}ulti-view {FPN} with position-aware attention for deep
  universal lesion detection,''
\newblock in {\em MICCAI}. Springer, 2019, pp. 13--21.

\bibitem{yan2019mulan}
Ke~Yan, Youbao Tang, Yifan Peng, Veit Sandfort, Mohammadhadi Bagheri, Zhiyong
  Lu, and Ronald~M. Summers,
\newblock ``{MULAN}: {M}ultitask universal lesion analysis network for joint
  lesion detection, tagging, and segmentation,''
\newblock in {\em MICCAI}. Springer, 2019, pp. 194--202.

\bibitem{tang2019uldor}
Youbao Tang, Ke~Yan, Yuxing Tang, Jiamin Liu, Jing Xiao, and Ronald~M. Summers,
\newblock ``{ULDor}: {A} universal lesion detector for {CT} scans with pseudo
  masks and hard negative example mining,''
\newblock {\em arXiv preprint arXiv:1901.06359}, 2019.

\bibitem{he2017mask}
Kaiming He, Georgia Gkioxari, Piotr Doll{\'a}r, and Ross Girshick,
\newblock ``Mask {R-CNN},''
\newblock in {\em ICCV}, 2017, pp. 2961--2969.

\bibitem{zlocha2019improving}
Martin Zlocha, Qi~Dou, and Ben Glocker,
\newblock ``Improving {RetinaNet} for {CT} lesion detection with dense masks
  from weak {RECIST} labels,''
\newblock {\em arXiv preprint arXiv:1906.02283}, 2019.

\bibitem{rother2004grabcut}
Carsten Rother, Vladimir Kolmogorov, and Andrew Blake,
\newblock ``{GrabCut}: {I}nteractive foreground extraction using iterated graph
  cuts,''
\newblock in {\em ACM Trans. Graphics}. ACM, 2004, vol.~23, pp. 309--314.
\newpage

\bibitem{tang2018semi}
Youbao Tang, Adam~P. Harrison, Mohammadhadi Bagheri, Jing Xiao, and Ronald~M.
  Summers,
\newblock ``Semi-automatic {RECIST} labeling on {CT} scans with cascaded
  convolutional neural networks,''
\newblock in {\em MICCAI}. Springer, 2018, pp. 405--413.

\bibitem{zhou2019bottom}
Xingyi Zhou, Jiacheng Zhuo, and Philipp Krahenbuhl,
\newblock ``Bottom-up object detection by grouping extreme and center points,''
\newblock in {\em CVPR}, 2019, pp. 850--859.

\bibitem{newell2016stacked}
Alejandro Newell, Kaiyu Yang, and Jia Deng,
\newblock ``Stacked hourglass networks for human pose estimation,''
\newblock in {\em ECCV}. Springer, 2016, pp. 483--499.

\bibitem{law2018cornernet}
Hei Law and Jia Deng,
\newblock ``{CornerNet}: {Detecting} objects as paired keypoints,''
\newblock in {\em ECCV}, 2018, pp. 734--750.

\bibitem{wang2019towards}
Xudong Wang, Zhaowei Cai, Dashan Gao, and Nuno Vasconcelos,
\newblock ``Towards universal object detection by domain attention,''
\newblock in {\em CVPR}, 2019, pp. 7289--7298.

\bibitem{wang2019volumetric}
Xudong Wang, Shizhong Han, Yunqiang Chen, Dashan Gao, and Nuno Vasconcelos,
\newblock ``Volumetric attention for {3D} medical image segmentation and
  detection,''
\newblock in {\em MICCAI}. Springer, 2019, pp. 175--184.

\bibitem{zhang20193d}
Ning Zhang, Dechun Wang, Xinzi Sun, Pengfei Zhang, Chenxi Zhang, Yu~Cao, and
  Benyuan Liu,
\newblock ``{3D} anchor-free lesion detector on computed tomography scans,''
\newblock {\em arXiv preprint arXiv:1908.11324}, 2019.

\end{thebibliography}

\end{document}